\documentclass[conference, compsoc]{IEEEtran}

\usepackage{lipsum}
\usepackage{times}
\usepackage{epsfig}
\usepackage{graphicx}
\usepackage{amsmath}
\usepackage{amssymb}
\usepackage{subcaption}
\usepackage{stfloats}
\usepackage{mathtools}
\usepackage{amsfonts}
\usepackage{ragged2e}
\usepackage{booktabs, multirow, tabularx}
\newcolumntype{L}{>{\RaggedRight}X}

\usepackage[pagebackref=true,breaklinks=true,letterpaper=true,colorlinks,bookmarks=false]{hyperref}

\newcommand{\by}{{\mkern-2mu\times\mkern-2mu}}
\newcommand{\SO}[1]{\ensuremath{\textrm{SO}(#1)}}
\newcommand{\SE}[1]{\ensuremath{\textrm{SE}(#1)}}
\newcommand{\SIM}[1]{\ensuremath{\textrm{SIM}(#1)}}
\newcommand{\SA}[1]{\ensuremath{\textrm{SA}(#1)}}
\newcommand{\GA}[1]{\ensuremath{\textrm{GA}^{+}(#1)}}

\newcommand{\so}[1]{\ensuremath{\mathfrak{so}(#1)}}

\newcommand{\sa}[1]{\ensuremath{\mathfrak{sa}(#1)}}
\newcommand{\ga}[1]{\ensuremath{\mathfrak{ga}^{+}(#1)}}

\newcommand{\norm}[1]{\ensuremath{\left \Vert #1 \right \Vert}}
\newcommand{\abs}[1]{\ensuremath{\left \vert #1 \right \vert}}

\def\eg{\emph{e.g. }} 
\def\ie{\emph{i.e. }} 
\def\etal{\emph{et al. }}

\makeatletter
\newcommand\subparagraph{%
  \@startsection{subparagraph}{5}
  {\parindent}
  {3.25ex \@plus 1ex \@minus .2ex}
  {-1em}
  {\normalfont\normalsize\bfseries}}
\makeatother
\usepackage{titlesec}
\let\subparagraph\relax

\titlespacing\section{0pt}{10pt plus 2pt minus 2pt}{7pt} 
\titlespacing\subsection{0pt}{8pt plus 2pt minus 2pt}{6pt}

\begin{document}

\title{Quotienting Impertinent Camera Kinematics for 3D Video Stabilization}
\author{
\begin{tabular}[t]{c@{\extracolsep{2em}}c@{\extracolsep{2em}}c} 
Thomas W. Mitchel & Christian W\"ulker & Jin Seob Kim \\
{\tt\small tmitchel@jhu.edu} & {\tt\small christian.wuelker@jhu.edu} & {\tt\small jkim115@jhu.edu} 
\end{tabular} \\ [4pt]
\begin{tabular}[t]{c@{\extracolsep{2em}}c} 
Sipu Ruan & Gregory S. Chirikjian \\
{\tt\small ruansp@jhu.edu} & {\tt\small gchirik1@jhu.edu}
\end{tabular} \\[4pt]

Laboratory for Computational Sensing and Robotics \\
Johns Hopkins University
}
\maketitle

\begin{abstract}
 With the recent advent of methods that allow for real-time computation, dense 3D flows have become a viable basis for fast camera motion estimation. Most importantly, dense flows are more robust than the sparse feature matching techniques used by existing 3D stabilization methods, able to better handle large camera displacements and occlusions similar to those often found in consumer videos. Here we introduce a framework for 3D video stabilization that relies on dense scene flow alone. The foundation of this approach is a novel camera motion model that allows for real-world camera poses to be recovered directly from 3D motion fields. Moreover, this model can be extended to describe certain types of non-rigid artifacts that are commonly found in videos, such as those resulting from zooms. This framework gives rise to several robust regimes that produce high-quality stabilization of the kind achieved by prior full 3D methods while avoiding the fragility typically present in feature-based approaches. As an added benefit, our framework is fast: the simplicity of our motion model and efficient flow calculations combine to enable stabilization at a high frame rate. 
\end{abstract}

\section{Introduction}
\begin{figure}[!t] 
\centering
\begin{subfigure}{.48\linewidth}
\includegraphics[width=.89\textwidth]{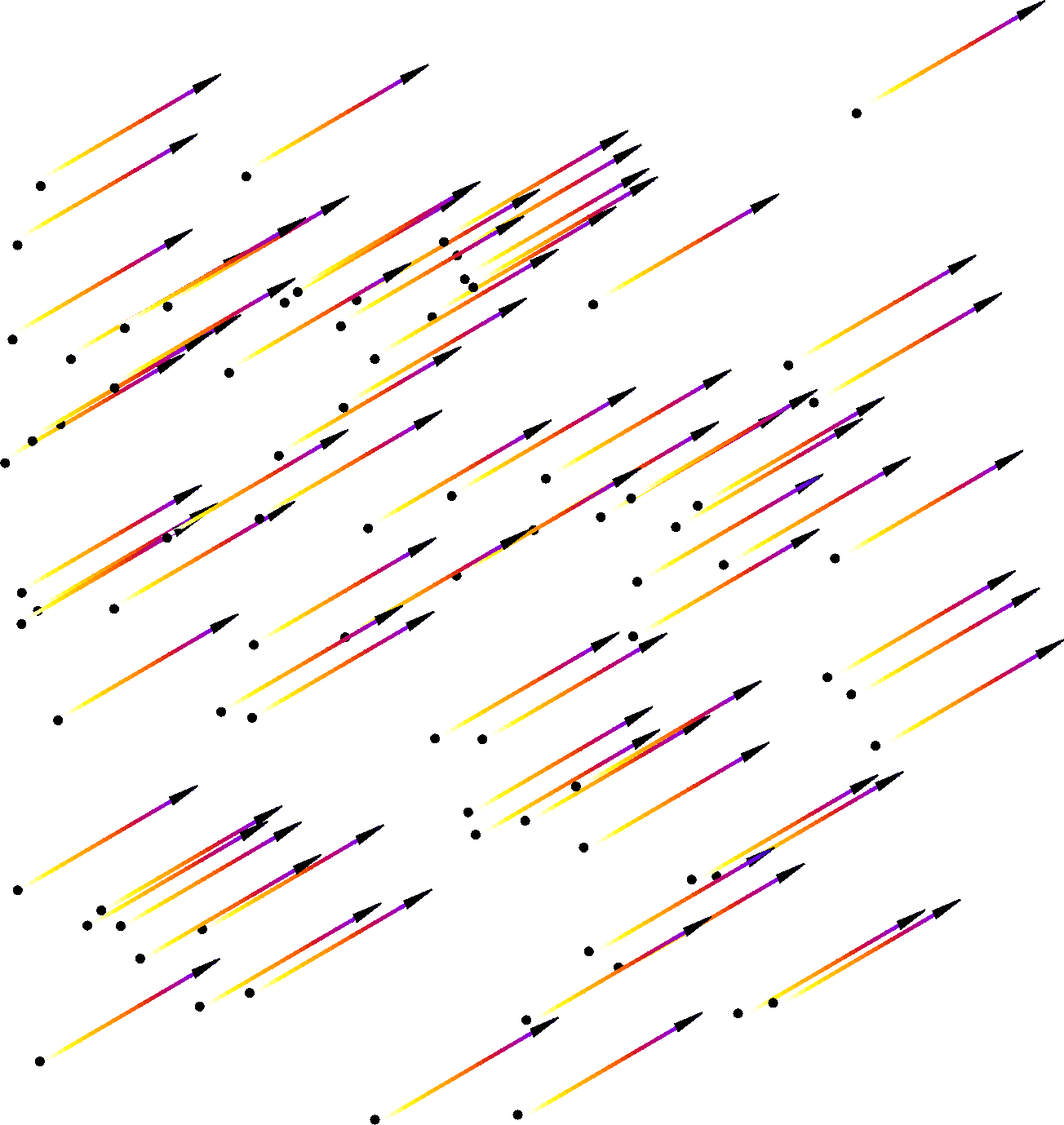}
\caption{Translation}
\label{fig:translation}
\end{subfigure}
\hfill
\begin{subfigure}{.48\linewidth}
\includegraphics[width=\textwidth]{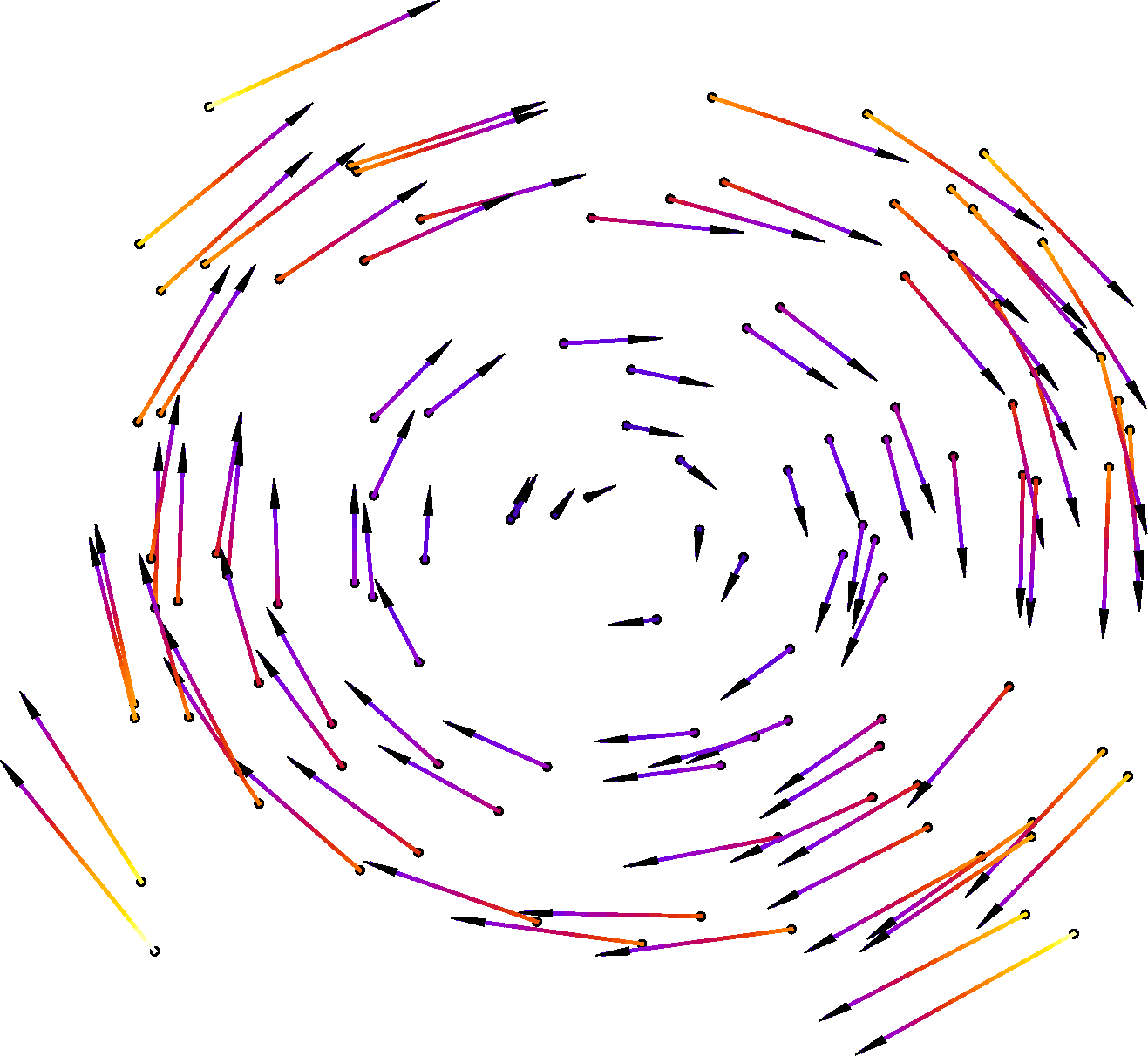}
\caption{Rotation}
\label{fig:rot}
\end{subfigure}
\vskip \baselineskip
\begin{subfigure}{.48\linewidth}
\includegraphics[width=\textwidth]{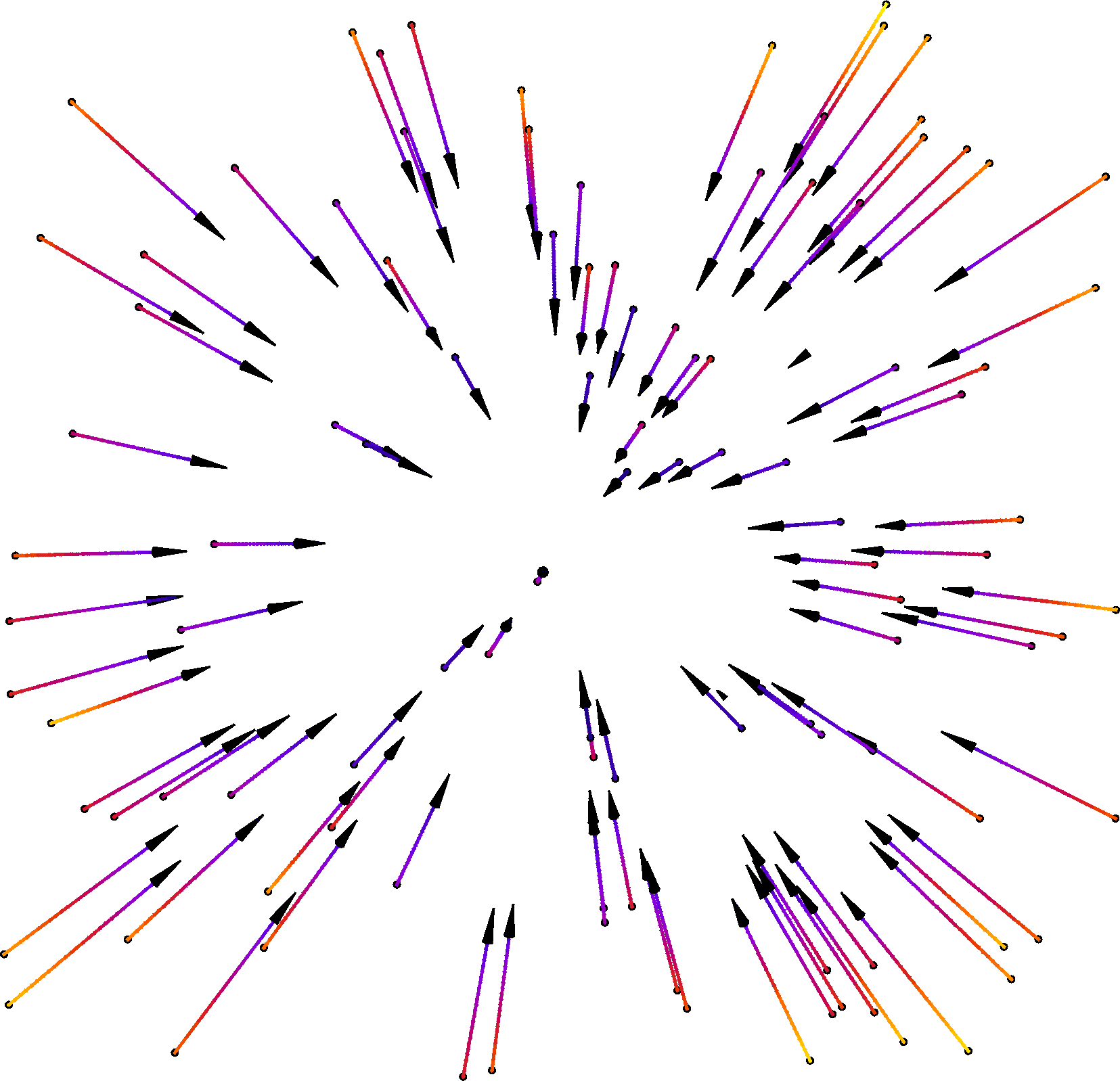}
\caption{Zoom}
\label{fig::scale}
\end{subfigure}
\hfill
\begin{subfigure}{.48\linewidth}
\includegraphics[width=\textwidth]{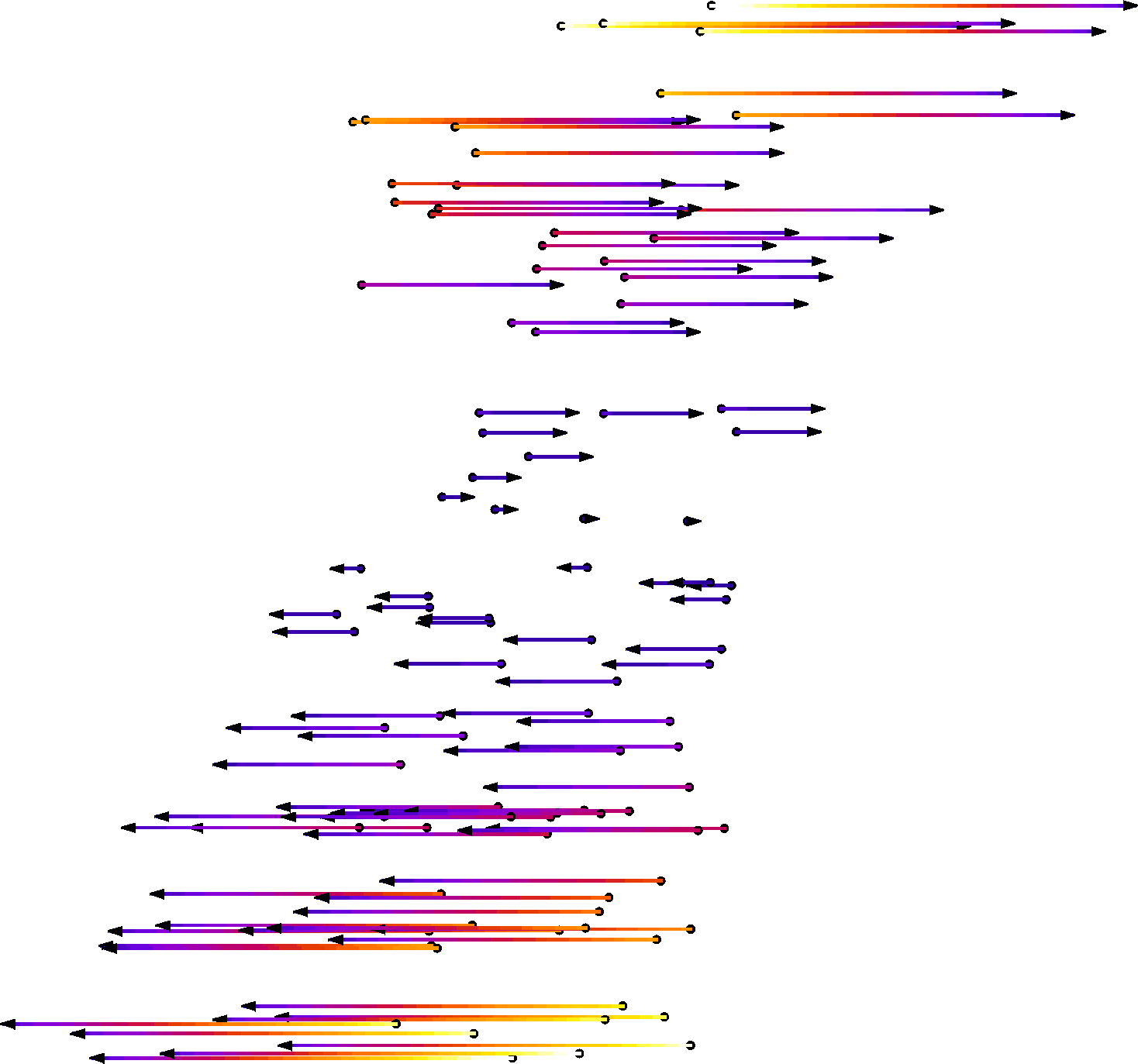}
\caption{Shear-like distortion effect}
\label{fig::shear}
\end{subfigure}
\caption{`Signatures' of different camera motion artifacts in flows as seen from the image plane}
\label{fig:motionf}
\end{figure}

Video stabilization algorithms depend on some measure of the original camera motion to remove unwanted artifacts and produce smooth videos.  Most existing 2D and 3D stabilization techniques recover inter-frame camera motion by either tracking feature trajectories \cite{grundmann2011auto, liu2009content, liu2011subspace} or matching image features between frames \cite{matsushita2006full, liu2012video, liu2013bundled}. Image features are  brittle and break down in videos with fast camera motions or scene occlusions, both of which are common in consumer videos. In contrast, dense flows (such as 2D optical flow or 3D scene flow) provide more robust descriptions of scene motion and remain accurate in challenging videos where feature-based approaches can fail \cite{sand2008particle, brox2010object, sundaram2010dense, wang2013dense}. As seen in Figure \ref{fig:motionf}, flows can contain information about both rigid camera motion and non-rigid artifacts, such as those produced by zooms or distortion effects. Some recent 2D stabilization methods \cite{liu2014steadyflow, liu2016meshflow, guo2018view} have incorporated the use of flows into camera motion estimation. However, each still rely on image features in some capacity and thus retain the weakness inherent in feature-based approaches. 

Our contribution is a technique for full 3D video stabilization that relies \textit{only} on dense scene flow and as a result, is able to take full advantage of the benefits provided by dense flows. Our approach is based on a novel motion model which allows for the recovery of 3D camera motions from dense scene flow alone.  Camera motion is represented by global transformations in a Lie group, and we exploit a powerful geometric relationship between 3D motion fields and twists to recover camera poses from scene flow via a closed-form expression. This model gives rise to a  natural formulation of the optimal camera path, and together they form a robust framework for 3D video stabilization which we call \textit{Quotienting Impertinent Camera Kinematics} or \textbf{QuICK} for short. 

 QuICK models camera motion by the action of 3D transformations on the real-world camera pose.  These transformations belong to the Lie groups \SE{3} (rigid-body motion), \SIM{3} (similarities), \SA{3} (special affine transformations -- analogous to homographies and translations), and \GA{3} (general affine transformations).  Fast scene flow \cite{jaimez2015primal} is used to produce the 3D motion field from which the transformations describing inter-frame motion are extracted. The transformations are then used to compute an optimal camera path that minimizes inter-frame camera motion subject to desired end-constraints. This framework results in robust video stabilization regimes which achieve high-quality stabilization while avoiding many of the pitfalls commonly encountered by other 3D methods.  QuICK is quick. The use of fast scene flow compliments QuICK's closed-form solution to the camera motion, enabling stabilization at a high frame-rate.  To the authors knowledge, QuICK is the first 3D video stabilization technique that does not rely on image features in any capacity. This makes QuICK  exceptionally flexible, able to handle challenging camera motions such as rapid rotations and zooms in videos where other 3D stabilization methods typically fail.

\section{Related work}

Existing video stabilization methods can be categorized as 2D, 2.5D, and 3D regimes.  Most 2D methods rely exclusively on matching image features \cite{shi1993good} between consecutive frames. Matsushita \etal \cite{matsushita2006full} represented the camera path through a series of homographies. Grundmann \etal \cite{grundmann2011auto} constructed cinematographic \cite{gleicher2007re} L1 optimal camera paths piecewise, utilizing a combination of similarity and homography transformations. Liu \etal \cite{liu2013bundled} introduced a non-linear image warping motion model used by many recent 2D methods. Recently, a series of 2D stabilization methods have been developed that utilize 2D motion fields to recover camera motions. Liu \etal \cite{liu2014steadyflow} developed the first such method in which feature tracking and dense optical flows were used to produce smoothed motion fields. Liu \etal \cite{liu2016meshflow} and Guo \etal \cite{guo2018view} then produced similar regimes for monocular and stereoscopic stabilization, respectively.  Liu \etal also developed a `codingflow' regime, using a 2D motion field comprised of smoothed motion vectors employed in video coding.  With the exception of `codingflow', all of these methods still rely on image features to some degree. 

Virtually all stabilization methods that incorporate 3D information use feature-based approaches. Long feature tracks represent a middle ground between 2D and 3D scene descriptions.  Liu and Gleicher \cite{liu2011subspace} exploited the partial 3D information contained in smoothed 2D feature trajectories to develop the first 2.5D method. Subsequently, both Goldstein and Fattal \cite{goldstein2012video} and Wang \etal \cite{wang2013spatially} introduced methods based on improving the length and quality of tracked feature trajectories. Recently, Liu \etal \cite{liu2017hybrid} developed a hybrid 2.5D regime which combines warping and homography motion models using both 2D and 3D features.

3D stabilization involves the use of 3D scene information to estimate the change in the 3D pose of the camera between frames. Liu \etal \cite{liu2009content} developed the first full 3D stabilization regime by smoothing trajectories of 3D points recovered using structure-from-motion (SFM) \cite{hartley2003multiple} and guiding `content preserving' image warps with their projections. Other successful 3D methods rely on supplemental cameras to avoid the need for brittle scene reconstruction. Smith \etal \cite{smith2009light} and Liu \etal \cite{liu2012video} used light field and depth cameras, respectively, to project matched features into 3D and recover the 3D camera pose. However, by relying on sparse feature matches these regimes remain sensitive to occlusions, dynamic scene content, and fast camera motions under which feature matching can break down. 

In contrast to prior 3D stabilization methods, QuICK uses dense scene flow alone to recover 3D camera poses. The result is a framework that produces high-quality stabilization without the accompanying fragility present in prior 3D regimes.

\section{The QuICK motion model}

QuICK is built upon a novel method for recovering 3D camera motion directly from dense scene flow. To motivate our approach, we begin by discussing the advantages of dense flows over image features in the context of video stabilization. Then we introduce the QuICK motion model and describe how 3D camera motion can be estimated from dense flows. Though we demonstrate our method with RGB-D videos, we are convinced that our motion model can be applied to any type of videos where 3D scene structure can be estimated and dense scene flow computed, such as stereoscopic or light field video. 

\subsection{Dense scene flow for 3D motion estimation}
In contrast to sparse flows, which are calculated by matching features, dense flows rely on a variational framework to give a complete description of the motion field covering an image. The most important consequence of this approach is that dense flows are much more robust than feature matches and by extension, sparse flows. They remain more accurate over long videos and better handle large camera displacements and occlusions \cite{sand2008particle, brox2010object, sundaram2010dense, wang2013dense}. This is of critical importance in video stabilization as many consumer videos are captured by cameras embedded in hand-held devices such as smart phones. 

Typically, feature matches and sparse flows have been cheaper to compute than dense flows. However, recent techniques \cite{kroeger2016fast, Sun_2018_CVPR} have made the estimation of complete motion fields significantly more efficient, making them a viable basis for fast motion estimation and video stabilization. Modern GPUs have helped to extend this trend to dense 3D flows, which we consider here. In particular, we use PD-Flow \cite{jaimez2015primal} for fast scene flow calculation from RGB-D intensity and depth image pairs. These developments have not been restricted to RGB-D videos, and other methods exist for fast scene flow calculation with stereoscopic videos \cite{taniai2017fast}. 

\subsection{Rigid-body motion: \SE{3}}
\begin{figure}[!t] 
\centering
\includegraphics[width=.45\textwidth]{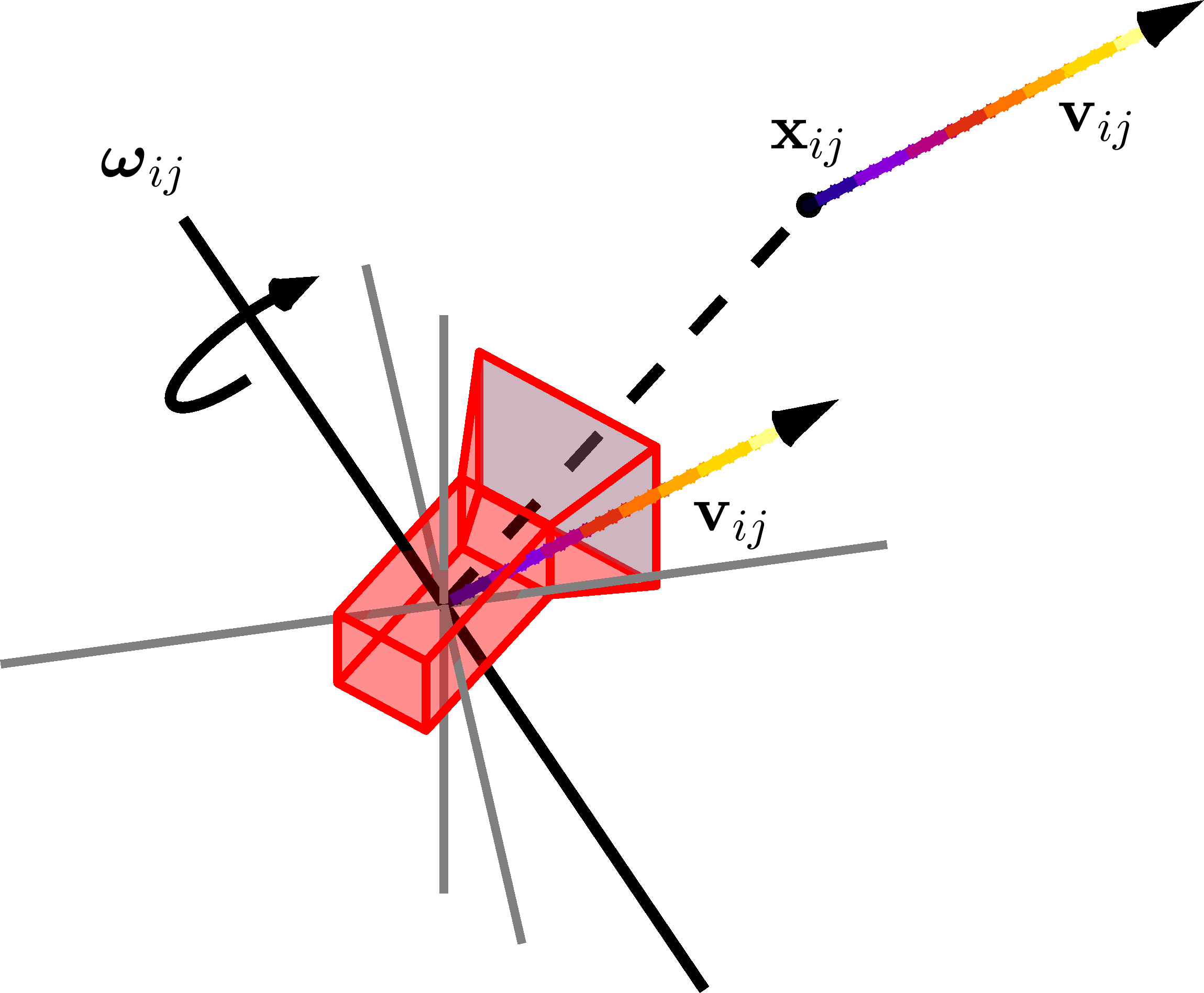}
\caption{The geometry of the induced twist $\xi_{ij}$ at the camera frame corresponding to the point ${\bf x}_{ij}$ with velocity ${\bf v}_{ij}$.}
\label{fig:induced_twist}
\end{figure}

\begin{figure*}[!t] 
\centering
\begin{subfigure}{.32\linewidth}
\includegraphics[width=.92\textwidth]{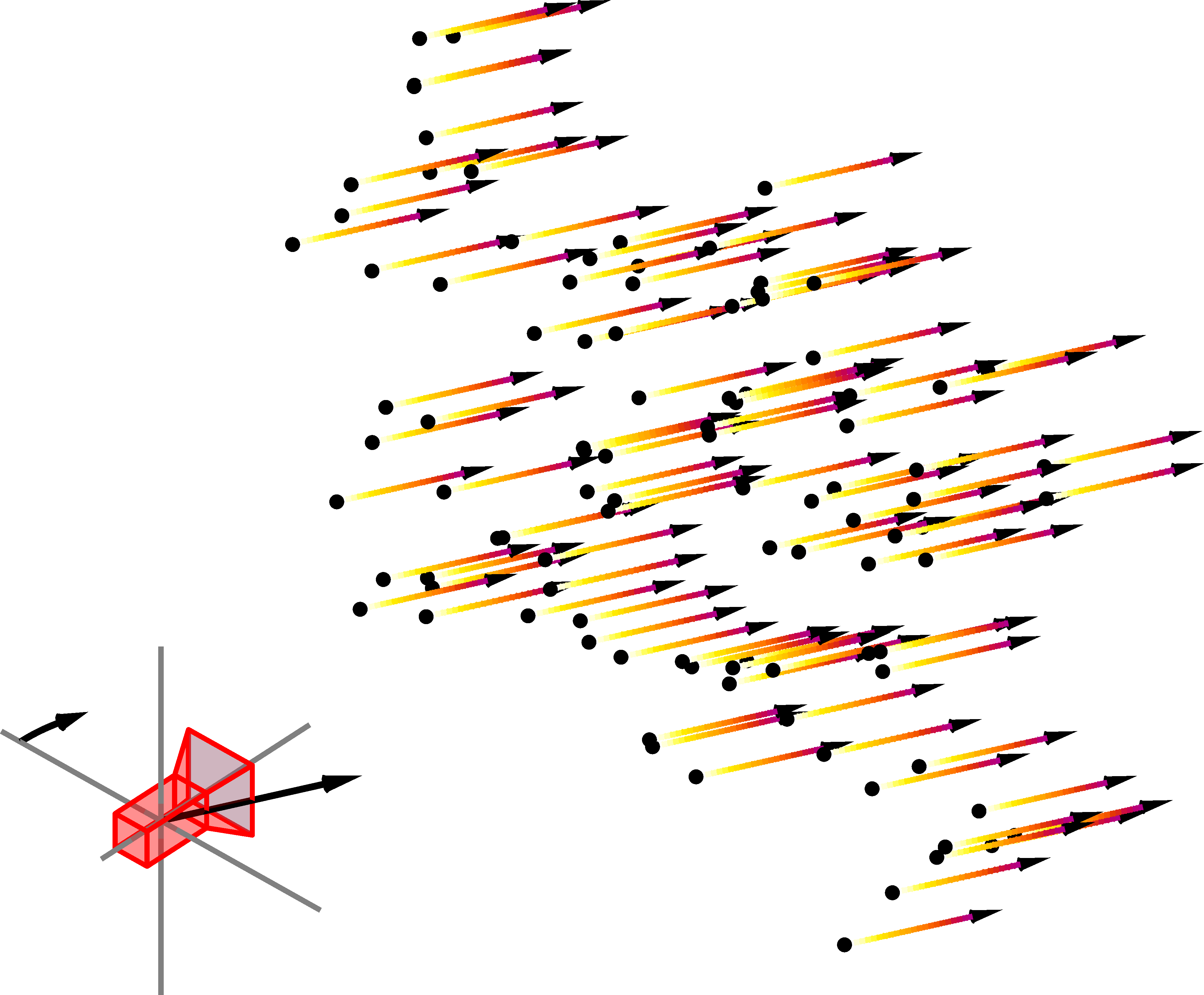}
\caption{Translational motion}
\label{fig:cam_trans}
\end{subfigure}
\hfill
\begin{subfigure}{.32\linewidth}
\includegraphics[width=.81\textwidth]{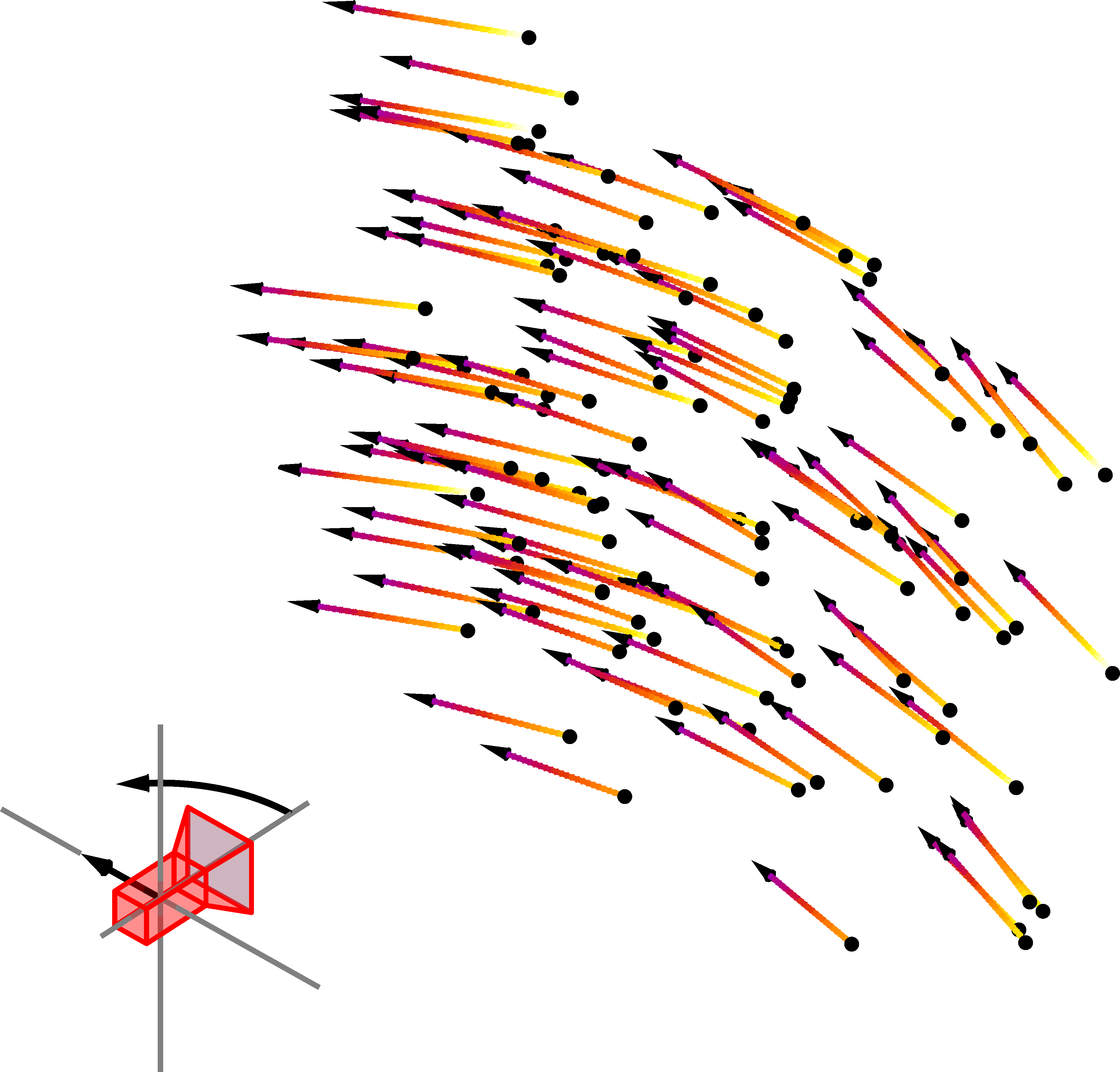}
\caption{Rotational motion: vertical axis}
\label{fig:cam_rot_z}
\end{subfigure}
\hfill
\begin{subfigure}{.32\linewidth}
\includegraphics[width=.89\textwidth]{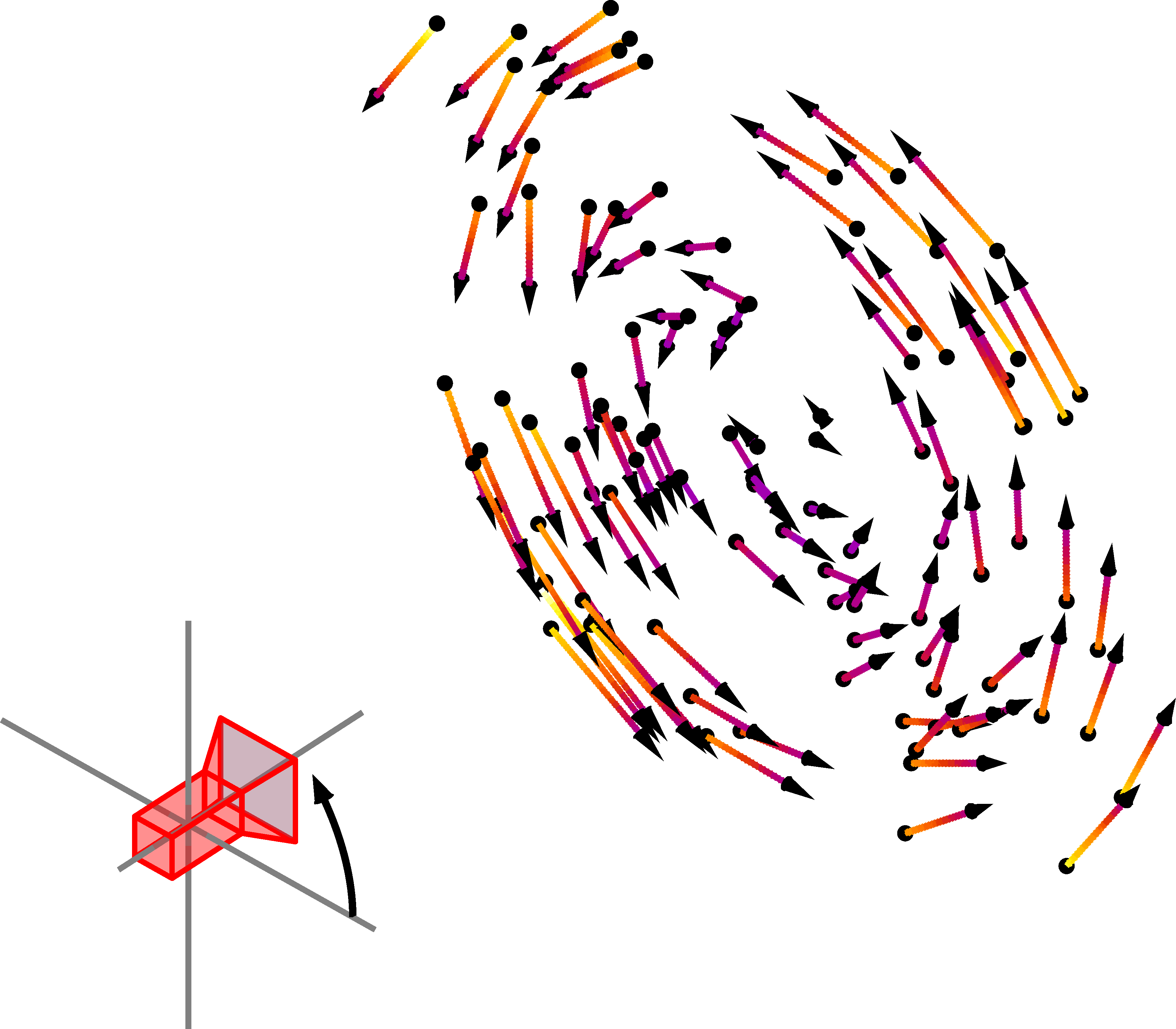}
\caption{Rotational motion: forward axis}
\label{fig:cam_rot_x}
\end{subfigure}
\caption{Examples of different rigid camera motions recovered from flows.}
\label{fig:camera_motion}
\end{figure*}
3D camera motion can be described by global parametric transformations belonging to a matrix Lie group. We begin by considering camera motion consisting of 3D translations and rotations, \ie rigid-body motions. These motions belong to the Lie group \SE{3}, which is the semi-direct product of $\mathbb{R}^3$ with the group of orientation-preserving 3D rotations, \SO{3}, \ie $\SE{3} = \mathbb{R}^3 \rtimes \SO{3}$. 

Suppose we are given a sequence of intensity images, their corresponding depth maps, and that the camera is in motion throughout the video. Assuming the camera calibration is known, all pixels in a given frame with valid depth information can be mapped to their real-world 3D points as seen from the camera frame, namely ${\bf x} \in \mathbb{R}^3$. Computing the scene flow gives the 3D translational velocity of each point ${\bf v} \in \mathbb{R}^3$. Now, let us consider an individual point, say the one corresponding to the $ij^{th}$ pixel, ${\bf x}_{ij}$, and imagine it to be rigidly fixed to the origin of the camera frame as shown in Figure~\ref{fig:induced_twist}. Since the two are fixed, the origin must undergo the same translational displacement. However, the displacement of the point also produces a corresponding \textit{angular} velocity at the origin, given by
\begin{equation}
\boldsymbol{\omega}_{ij} = \frac{{\bf x}_{ij} \times {\bf v}_{ij}}{{\bf x}_{ij}^T {\bf x}_{ij}}. \label{w}
\end{equation}
Together, they form a twist at the camera frame, \begin{equation} \label{xi_ij}
    \xi_{ij} = \begin{bmatrix} {\bf v}_{ij} \\ \boldsymbol{\omega}_{ij}  \end{bmatrix} \in \mathbb{R}^6,
\end{equation} which we call the \textit{induced twist}.

Applying this process to all of the 3D points, the twist describing the motion of the camera between frames, $\xi \in \mathbb{R}^6$, minimizes \begin{equation} f(\xi) = \frac{1}{2} \sum_{ij} \norm{\xi - \xi_{ij}}^2, \label{se3_cost} \end{equation}
where the sum is over all pixel indices with valid depth. Letting $D$ be the number of all such pixel indices, $\xi$ is simply the average of all induced twists, \begin{equation} \label{cam_twist}
    \xi = \frac{1}{D} \sum_{ij}\xi_{ij} = \begin{bmatrix} \bf{v} \\ \boldsymbol{\omega} \end{bmatrix},
\end{equation} which we call the \textit{camera twist}.

Regardless of frame rate, we can interpret the camera twist $\xi$ as the \textit{displacement} of the camera between frames, such that $\bf{v}$ is the translation of the camera and $\boldsymbol{\omega}$ is the axis of rotation in \so{3}, the Lie algebra of \SO{3}.  Together, they are the twist coordinates  that parameterize the transformation $g \in \SE{3}$ representing the camera motion. Under this observation, we break from the classical structure of $\SE{3}$ and consider translational and rotational motion to be decoupled. This allows allows for the camera motion to be expressed by the homogeneous transformation \cite{chirikjian2018pose} \begin{equation}
    g\left(\xi\right) = \begin{bmatrix} \exp(\widehat{\boldsymbol{\omega}}) & {\bf v} \\ {\bf 0}^T & 1\end{bmatrix}, \label{se3_exp}
\end{equation} where $\widehat{\ \cdot \ }$ is defined such that for any ${\bf a}, {\bf b} \in \mathbb{R}^3$, $\widehat{\bf a} {\bf b} = {\bf a} \times {\bf b}$.  We find that only basic filtering of the flow by excluding points with unnaturally large velocities is required for robust estimation of the camera motion.

A visualization of the recovery of different camera motions in shown in Figure~\ref{fig:camera_motion}. It is interesting to note that due to the position of the scene in front of the camera, pure translations and rotations are not always recovered as such --- demonstrated by the small rotations and translations recovered in \ref{fig:cam_trans} and \ref{fig:cam_rot_z}, respectively. Scene symmetry exists only about and along the forward axis and corresponding motions can be recovered exactly as seen in \ref{fig:cam_rot_x}. Regardless, we find this has a negligible effect on stabilization. If the recovery of exact motions is desired, the induced twists can be computed at the geometrical center of the scene then transformed back to the camera frame.

\subsection{Similarities: \SIM{3}}
 Zooms are a common feature in consumer videos.  QuICK's framework can be extended to model them by considering camera motion in the Lie group \SIM{3}, consisting of rigid motions and isotropic scaling. The induced twist corresponding to an arbitrary point ${\bf x}_{ij}$ with displacement ${\bf v}_{ij}$ is of the form  \begin{equation}
    \xi_{ij} = \begin{bmatrix} {\bf v}_{ij} \\ \boldsymbol{\omega}_{ij} \\ a_{ij} \end{bmatrix}, \label{sim_ij}
\end{equation} the key insight being that the scaling component is given by   \begin{equation}
    a_{ij} = \frac{{\bf x}_{ij}^T {\bf v}_{ij}}{{\bf x}_{ij}^T {\bf x}_{ij}}. \label{sc_ij}
    \end{equation}
 Again, the camera twist, $\xi = \left[{\bf v}^T, \ \boldsymbol{\omega}^T, \ a \right]^T,$ is simply the average of all induced twists. Decoupling the translational component and exponentiating the rotational and scaling components gives the homogeneous transformation describing the camera motion, \begin{equation}
    g\left(\xi\right) = \begin{bmatrix} e^a \exp\left(\widehat{\boldsymbol{\omega}}\right) & {\bf v} \\ {\bf 0}^T & 1 \end{bmatrix}.
\end{equation} To our knowledge, our \SIM{3} implementation of QuICK is the first full 3D video stabilization regime to explicitly consider the effects of zooms.

\subsection{Affine transformations: \SA{3} and \GA{3}}

Modeling camera motion using higher dimensional Lie groups requires a general formulation of the QuICK motion model. In particular, the form of the induced twist corresponding to an arbitrary point depends on chosen basis for the Lie algebra. 

Let $G$ be an affine matrix Lie group (\ie the bottom row of each matrix element is filled  with zeros except for the last element, which is $1$) and suppose the matrices $E_1, \ \ldots, E_n$ form a basis for its corresponding Lie algebra, $\mathfrak{g}.$ Then, the induced twist at the camera frame corresponding to a point ${\bf x}_{ij}$ with velocity ${\bf v}_{ij}$ is of the form \begin{equation}
    \xi_{ij} = \Lambda_{ij}^{-T} \left[ \begin{smallmatrix} {\bf v}_{ij} \\ {\bf 0} \end{smallmatrix}\right] \in \mathbb{R}^n, \label{induced_twist}
\end{equation}
where $\Lambda_{ij} \in \mathbb{R}^{n \by n}$ is defined with respect to ${\bf x}_{ij}$ such that \begin{gather}
    \Lambda_{ij} = \left[\left(h_{ij}E_1 h_{ij}^{-1}\right)^{\vee}, \ \ldots, \ \left(h_{ij}E_n h_{ij}^{-1}\right)^{\vee}\right] \label{Lambda}, \\
    h_{ij} = \begin{bmatrix} \mathbb{I} & \frac{{\bf x}_{ij}}{{\bf x}_{ij}^T {\bf x}_{ij}} \\ {\bf 0}^T & 1 \end{bmatrix}. 
\end{gather} Here the $\vee$ operator is defined by an identification of Lie algebra basis elements, $\left\{ E_i \right \}$ with the natural unit basis vectors in $\mathbb{R}^n$, $\left \{{\bf e}_i \right \}$, together with the condition of linearity, \eg \begin{equation}
    \left(\sum_{i = 0}^n x_i E_i \right)^{\vee} = \sum_{i=0}^{n} x_i {\bf e}_i. 
\end{equation} The camera twist is the average of all induced twists, as in ~(\ref{cam_twist}).

The matrix $\Lambda_{ij}$ is a particular form of the matrix representation of the Lie group's adjoint operator \cite{chirikjian2011stochastic}, namely $\Lambda_{ij} = \left[Ad\left(h_{ij}\right)\right]$. To this point, the form of the induced twist in~(\ref{induced_twist}) is similar to the transformation of a pure force between coordinate frames in the theory of wrenches \cite{Murray:1994:MIR:561828}. However, normalizing the translational component $h_{ij}$ by dividing by the square of its magnitude allows us to preserve velocities instead of forces and torques. 

The 3D special affine group, \SA{3}, is the semi-direct product of $\mathbb{R}^3$ with $\textrm{SL}(3)$, the group of all unit determinant matrices in $\mathbb{R}^{3 \by 3}.$ $\textrm{SL}(3)$ is of interest to the problem of video stabilization as it can be roughly thought of as the group of homographies with unit determinant \cite{ma2012invitation}. Therefore, we believe that \SA{3} is worth examining in the context of 3D stabilization. 

The Lie algebra of $\textrm{SL}(3)$ is $\mathfrak{sl}(3)$, which consists of all zero-trace matrices in $\mathbb{R}^{3\by3}$. A possible basis for $\mathfrak{sl}(3)$ is \begin{gather}
    \widetilde{E}_1 = \widehat{\bf e}_1, \ \widetilde{E}_2 = \widehat{\bf e}_2, \ \widetilde{E}_3 = \widehat{\bf e}_3, \\
    \widetilde{E}_4 = \abs{\widehat{\bf e}_1}, \ \widetilde{E}_5 = \abs{\widehat{\bf e}_2}, \ \widetilde{E}_6 = \abs{\widehat{\bf e}_3}, \\
    \widetilde{E}_7 = \begin{bmatrix} 1 & 0 & 0 \\ 0 & -1 & 0 \\ 0 & 0 & 0 \end{bmatrix}, \ \widetilde{E}_8 = \begin{bmatrix} 0 & 0 & 0 \\ 0 & 1 & 0 \\ 0 & 0 & -1 \end{bmatrix},
\end{gather} where $\abs{\ \cdot \ }$ denotes the absolute value of the matrix elements. We can choose a basis for $\sa{3}$ based on $\mathfrak{sl}(3)$, namely
\begin{gather}
   E_k = \begin{dcases}
 \ \begin{bmatrix} \mathbb{O}_3 & {\bf e}_k \\ {\bf 0}^T & 0 \end{bmatrix}, & 1 \leq k \leq 3 \\
 \begin{bmatrix} \widetilde{E}_{k-3} & {\bf 0} \\ {\bf 0}^T & 0 \end{bmatrix}, & 4 \leq k \leq 11
 \end{dcases}. \label{sa_basis}
\end{gather}
Expressed relative to this basis, the induced twist corresponding to an arbitrary point ${\bf x}_{ij}$ with displacement ${\bf v}_{ij}$ is of the form \begin{equation}
        \xi_{ij} = \begin{bmatrix} {\bf v}_{ij} \\ \boldsymbol{\omega}_{ij} \\  \frac{1}{{\bf x}_{ij}^T {\bf x}_{ij}}\abs{\widehat{\bf x}_{ij}} {\bf v}_{ij} \\ \frac{1}{{\bf x}_{ij}^T {\bf x}_{ij}}{\bf x}_{ij}^T \widetilde{E}_7 {\bf v}_{ij} \\
    \frac{1}{{\bf x}_{ij}^T {\bf x}_{ij}}{\bf x}_{ij}^T \widetilde{E}_8 {\bf v}_{ij} \end{bmatrix} \in \mathbb{R}^{11}. 
\end{equation}

All groups we have considered thus far are subgroups of the positive general affine group, \GA{3}, which is the semi-direct product of $\mathbb{R}^3$ with $\textrm{GL}^{+}(3)$, the group of all $3 \times 3$ matrices with positive determinant. As the most general 3D affine Lie group, it has the potential to best approximate 3D rigid camera motions and non-rigid artifacts via a single, global transformation.  A possible basis for the Lie algebra of $\textrm{GL}^{+}(3)$, $\mathfrak{gl}^{+}(3)$ is given by the matrices\begin{equation}
    \left[\widetilde{E}_{k}\right]_{mn} = \begin{dcases}
    1, & 3 (n - 1) + m = k \\
    0, & \textrm{otherwise}
    \end{dcases}, \ 1 \leq k \leq 9. 
\end{equation}
We can then define a basis for $\ga{3}$, the Lie algebra of $\GA{3}$, as in~(\ref{sa_basis}). In this basis, the form of the induced twist at the camera frame is \begin{equation}
    \xi_{ij} = \begin{bmatrix} 
    {\bf v}_{ij} \\ \frac{1}{{\bf x}_{ij}^T {\bf x}_{ij}}  {\bf x}_{ij}^T {\bf e}_1 {\bf v}_{ij} \\\frac{1}{{\bf x}_{ij}^T {\bf x}_{ij}}  {\bf x}_{ij}^T {\bf e}_2 {\bf v}_{ij} \\
   \frac{1}{{\bf x}_{ij}^T {\bf x}_{ij}}  {\bf x}_{ij}^T {\bf e}_3 {\bf v}_{ij} \end{bmatrix} \in \mathbb{R}^{12}. 
\end{equation}
The standard matrix exponential does not provide a surjective mapping from either the $\mathfrak{sl}(3)$ or $\mathfrak{gl}^{+}(3)$ to their corresponding Lie groups \cite{zacur2014left}. However, the mapping \begin{equation}
    \phi(U) = \exp\left(U^T\right) \exp\left(U - U^T\right) \label{phi}
\end{equation}
is surjective for both $\mathfrak{sl}(3)$ and $\mathfrak{gl}^{+}(3)$ \cite{andruchow2014left, vandereycken2013riemannian}. Assuming the general form of the camera twist to be $\xi = \left[{\bf v}^T, {\bf u}^T\right]^T$, the camera motion is \begin{equation}
g\left(\xi\right) = \begin{bmatrix} \phi \left(U\right) & {\bf v} \\ {\bf 0}^T & 1 \end{bmatrix},
\end{equation} where $U$ is the element in either $\mathfrak{sl}(3)$ or $\mathfrak{gl}^{+}(3)$ with twist coordinates ${\bf u}$. 

\section{Video stabilization}
\begin{figure}[!t] 
\centering
\includegraphics[width=.47\textwidth]{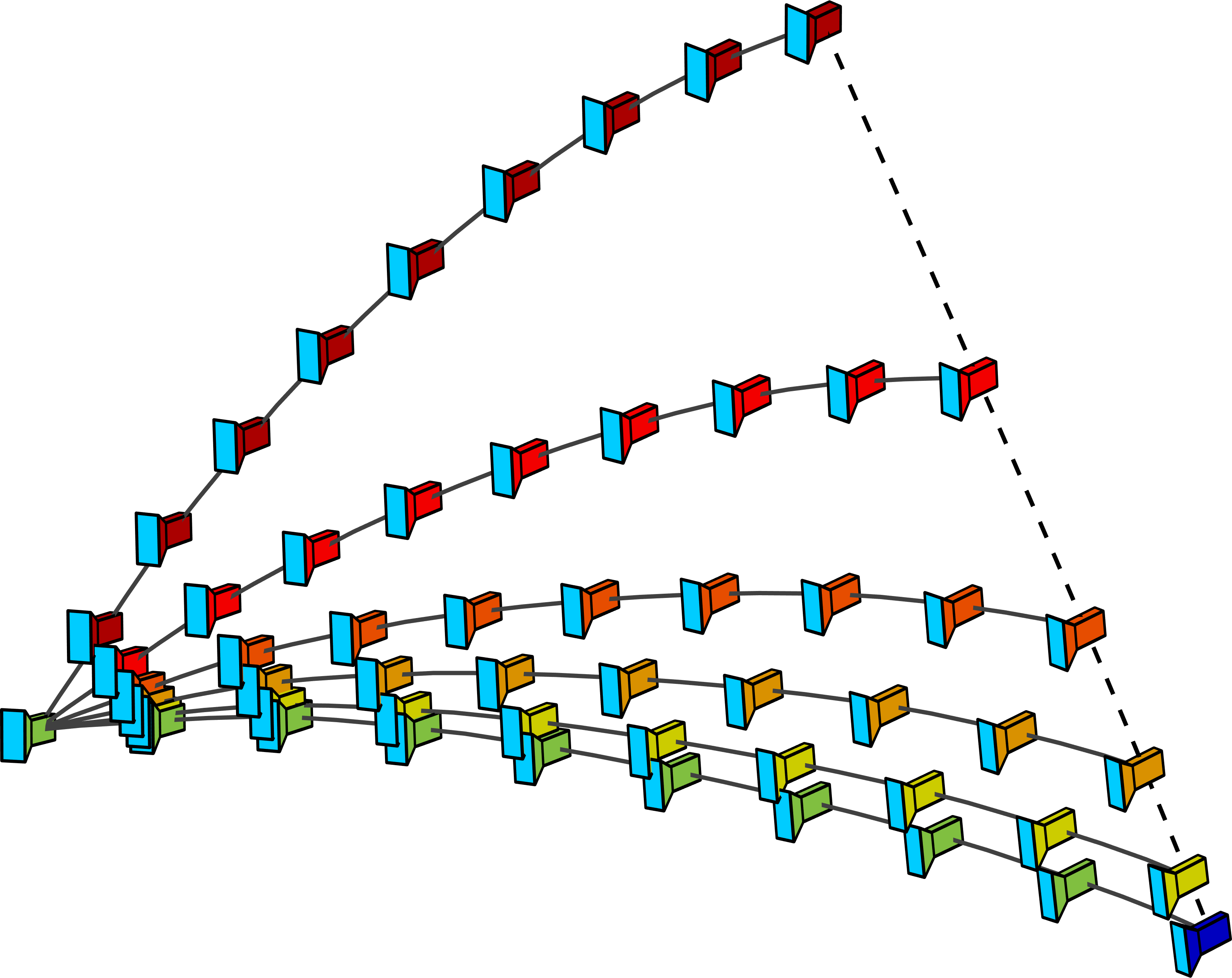}
\caption{The end-constraint solution process visualized with respect to the stabilized camera paths.}
\label{fig:ec_soln}
\end{figure}

When depth information is known, the QuICK motion model can be applied to estimate the 3D inter-frame camera motion over the course of a video in terms of the chosen affine Lie group, $G$. Taking $H$ to be the Lie group such that $G = \mathbb{R}^3 \rtimes H$, the result is the camera twist as a function of time, \begin{equation}
    \xi(t) = \begin{bmatrix} {\bf v}(t) \\ {\bf u}(t) \end{bmatrix} \in \mathbb{R}^n,
\end{equation} where ${\bf u}$ is a vector of Lie algebra twist coordinates corresponding to $H$. As $\xi$ measures relative displacement between frames, we integrate on the right to recover the estimated 3D camera path as seen from the initial camera pose, $g_{c}(t) \in G$, \eg \begin{gather}
g_c\left(0\right) = \mathbb{I}_4, \\
g_c\left(t_{i+1}\right) = g_c\left(t_i\right) g\left(\xi(t_{i})\right).
\end{gather} 

Conversely, $g_c^{-1}(t)$ maps the camera back to its initial pose. When camera travels only a small distance or is intended to remain static, $g_c^{-1}$ can be used to stabilize the video. However, the applications of this approach are limited.

\subsection{Optimal camera paths}
Let $\sigma$ be the mapping which takes the twist coordinates ${\bf u}$ to their corresponding element in the Lie group $H$, such that $\sigma\left({\bf u}(t)\right) = A(t) \in H$. To formulate a robust description of an optimal camera path, we consider the twists \begin{equation}
\overline{\zeta}(t) = \begin{bmatrix} - A^{-1} {\bf v} \\ \sigma^{-1}\left(A^{-1}\right) \end{bmatrix} \in \mathbb{R}^n.
\end{equation} If $G = \SE{3}$ or $G = \SIM{3}$, $\sigma^{-1}$ is computed via the matrix logarithm, \ie \begin{equation} \sigma^{-1}\left(A^{-1}\right) = \log^{\vee}\left(A^{-1}\right) = - {\bf u}. \end{equation}
However, if the group is $\SA{3}$ or $\GA{3}$, the inverse of ~(\ref{phi}) has no closed form expression and $\sigma^{-1}$ must be computed numerically using gradient descent \cite{zacur2014left}.  
\begin{figure}[!t] 
\centering
\includegraphics[width=.45\textwidth]{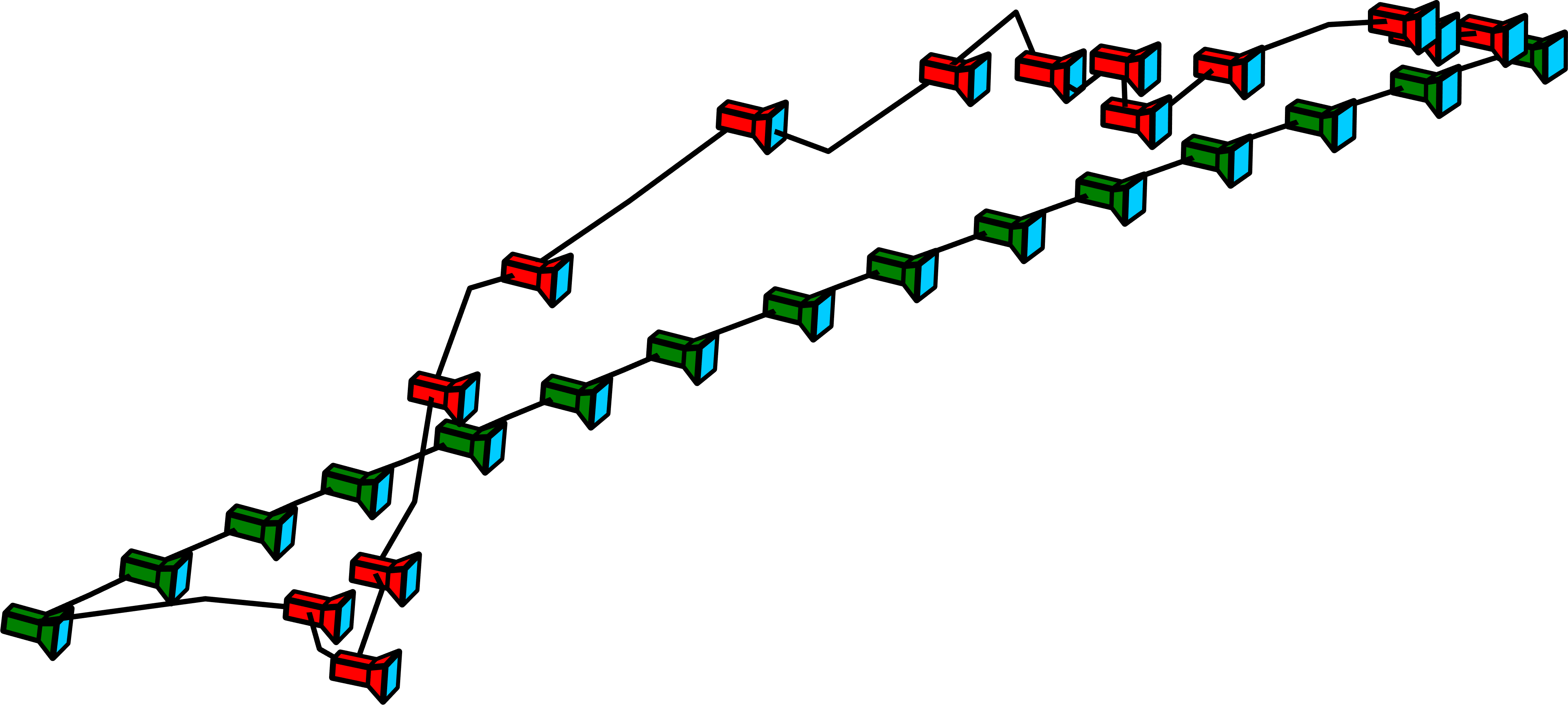}
\caption{An estimated camera path (red) and the corresponding optimal path (green).}
\label{fig:camera_path}
\end{figure}
The twists $\overline{\zeta}(t)$ correspond to the inverses of the motions parameterized by the camera twist $\xi(t)$. With this in mind, we consider the functional \begin{align}
    J\left[\zeta\right] &= \frac{1}{2}  \int_{t_0}^{t_f} \norm{\zeta(t) - \overline{\zeta}(t)}_{W}^2 dt \nonumber \\
    & = \frac{1}{2} \int_{t_0}^{t_f} \left[\zeta^T W \zeta - 2 {\bf b}^T \zeta + c\right] dt  \label{path},
\end{align} where $W = W^T \in \mathbb{R}^{n\by n}$ and \begin{gather}
{\bf b} = W \overline{\zeta} \\
c = \overline{\zeta}^T W \overline{\zeta}.
\end{gather} 
Subject to desired end-constraints, the twist that minimizes $J$, $\zeta^*(t)$, parameterizes the transformations that map the camera to the optimal path. Hence, QuICK gets its name --- imagining $S$ as the space on which videos temporally evolve, applying these transformations to the video reduces the `problem' of viewing the video from $S$ to the quotient space, $G \backslash S$. 

The variational minimization of~(\ref{path}) can be directly addressed by applying the coordinate-free generalization of the Euler-Lagrange equation known as the \textit{Euler-Poincar\'{e}} equation \cite{chirikjian2017signal}, through which the differential equation describing  the minimal solution can be recovered: \begin{equation}
\dot{\zeta} = W^{-1} \left[ \dot{{\bf b}} + \left[ad\left(\zeta\right) \right]^T \left[W \zeta - {\bf b}\right]\right]. \label{zeta_dot}
\end{equation} The matrix $\left[ad\left(\zeta\right)\right]$ is defined with respect to the chosen Lie algebra basis for $\mathfrak{g}$, $E_1, \ \ldots, \ E_n$, such that \begin{gather} 
\left[ad\left(\zeta\right)\right] = \left[\left(\left[\widehat{\zeta}, E_1\right]\right)^{\vee}, \ \ldots, \ \left(\left[\widehat{\zeta}, E_n\right]\right)^{\vee}\right],
\end{gather} where for any $A, B \in \mathbb{R}^{n\by n}$, $\left[A, B\right] = AB - BA.$ The matrix $W$ can be used to set the relative weights of different motions along the path, however, we do not explore this in detail and set $W = \mathbb{I}_n$ in our implementations.

Videos are stabilized piecewise. Key-frames are selected and the video is split into segments bounded by adjacent key-frames. In our implementations, key-frames are set at 30 frame intervals or when large changes in the direction of the velocity occur along the estimated camera path.  The formulation of~(\ref{path}) means that the chosen end-constraints for the path are in fact the transformations that map the estimated camera poses at the bounding key-frames to their desired, stabilized poses. If no corrections are desired at the key-frames, then the end-constraints are just the identity transformation. Since consecutive sequences share the same final and initial camera poses, the global camera path is smooth and continuous in the first derivative. 

We adapt the method of Kim and Chirikjian \cite{kim2006conformational} to solve~(\ref{zeta_dot}) subject to the desired end-constraints. While Kim and Chirikjian only consider $\SE{3}$, we find that the method easily generalizes to handle the other affine Lie groups we consider here.  An example of the solution process is shown in Figure~\ref{fig:ec_soln}. The differential equation is first integrated on the right with respect to an initial guess for the value of $ \zeta_0 = \zeta\left(0\right)$. Applying these transformations to the estimated camera path produces a stabilized path, shown in dark red. An artificial trajectory is defined between the distal correction transformation produced by integrating the guess and the desired correction transformation given by the end-constraints. In Figure~\ref{se3_cost}, this is visualized in terms of the camera path: the artificial trajectory, represented by the dashed line, is between the camera pose at the end of the initial path and the desired pose, highlighted in blue. The initial guess is then updated to track the trajectory \cite{han2001least} and the process is iterated until convergence. Figure~\ref{fig:camera_path} shows an example of an estimated camera trajectory (in red) and the corresponding stabilized path (in green).  

The end result of this process is a series of stabilizing transformations, $g_s(t) \in G$,  that map the camera to its stabilized path.  We accept a greater degree of global distortion in exchange for increased local image quality in and around occluded regions in the depth map and use `content-preserving' warps \cite{liu2009content} to synthesize stabilized frames. All pixels with valid depth form the control points. We use a $2 \times 2$ grid and set the relative weight between the data and similarity terms in the energy equation, $E = E_{d} + \alpha E_{s}$, to $\alpha = 0.1$.

\begin{table*}[t!]\centering 
\resizebox{\textwidth}{!}{\begin{tabular}{lcccccccc}
\multirow{2}[3]{*}{} & \multicolumn{4}{c}{2D} & \multicolumn{2}{c}{2.5D} & \multicolumn{2}{c}{3D} \\
\cmidrule(lr){2-5} \cmidrule(lr){6-7} \cmidrule(lr){8-9}
  & Bundled\cite{liu2013bundled} & Steady\cite{liu2014steadyflow} & Mesh\cite{liu2016meshflow} & Coding\cite{liu2017codingflow} & Epipolar\cite{goldstein2012video} & Hybrid\cite{liu2017hybrid} & Depth\cite{liu2012video} & \textbf{QuICK} \\
\midrule
 Speed (ms) & 392 & 1500 & 20 & 18 & 950 & 400 & 926 & 121 \\
\bottomrule
\end{tabular}}
\caption{Per-frame processing speed of different stabilization regimes.} \label{benchmark}
\end{table*}

\begin{figure*}[!t] 
\centering
\begin{subfigure}{\linewidth}
\includegraphics[width=\textwidth]{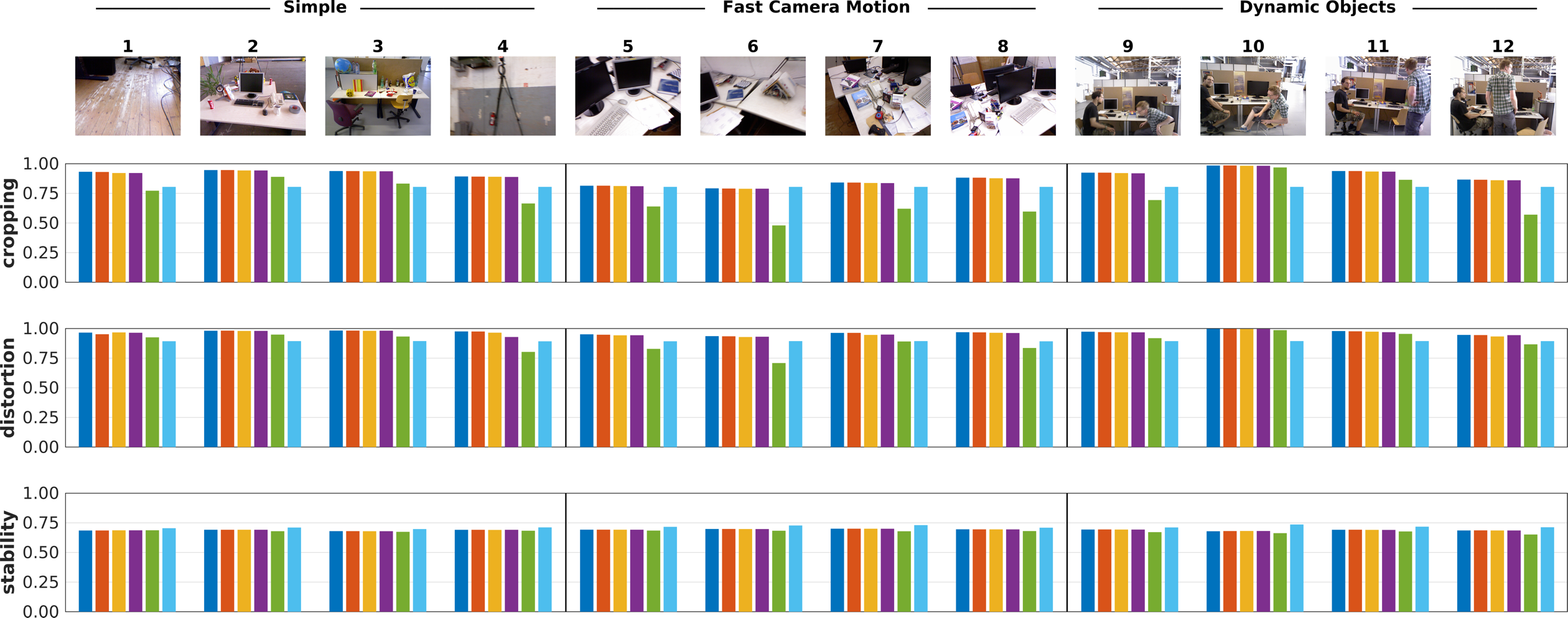}
\end{subfigure}
\vskip \baselineskip
\begin{subfigure}{\linewidth}
\centering
\includegraphics[width=.7\textwidth]{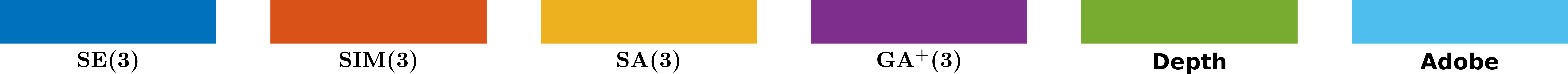}
\end{subfigure}
\caption{Quantitative comparisons using real videos from the TUM RGB-D SLAM dataset \cite{sturm12iros}.}
\label{fig:eval_slam}
\end{figure*}

\section{Experiments}

We run QuICK in C++ on a system with a 3.8 GHz CPU, 32GB of RAM, and an NVIDIA RTX 2080 GPU. Only scene flow calculations \cite{jaimez2015primal} are implemented on the GPU. For a 30-frame RGB-D sequence at 640 $\times$ 480 resolution, our \SE{3} implementation of QuICK takes approximately 121 ms to process a frame.  Specifically, computing the scene flow, recovering the camera motion, and computing the optimal path take 16 ms, 0.3 ms, and 4.2 ms, respectively. The main bottleneck is the rendering of the stabilized frame which takes 103 ms. For a 320 $\times$ 240 resolution sequence, we achieve a processing speed of 35 ms per frame (29 fps). Table~\ref{benchmark} shows the per-frame processing speed of other recent stabilization methods with videos of similar resolution. All data are from published results with the exception of \cite{liu2012video}, which we measure using our own implementation.  It is important to note that processing speed is dependent on the system on which a method is implemented. Generally, QuICK is significantly faster than existing 2.5D and 3D regimes and its processing speed is on par with that of successful 2D methods.

\subsection{Quantitative evaluation and comparisons}
To demonstrate the effectiveness of QuICK, we perform experiments using publicly available real and synthetic RGB-D sequences from the TUM RGB-D SLAM \cite{sturm12iros} and SceneNet RGB-D \cite{McCormac:etal:ICCV2017} datasets, respectively. We consider four different implementations of QuICK which model 3D camera motion using the Lie groups \SE{3}, \SIM{3}, \SA{3}, and \GA{3}, respectively. Examples of stabilized videos produced by each of these implementations can be found in the supplementary material. To evaluate the quality of stabilized videos, we adopt three objective quantitative metrics \cite{liu2013bundled}: \textit{cropping}, \textit{distortion}, and \textit{stability} defined as described in \cite{liu2017codingflow}. We are not aware of any publicly available implementations of prior 3D stabilization regimes, so we compare QuICK to our own implementation of the full 3D `depth camera' stabilization method \cite{liu2012video}.  Additionally, we compare against the `Warp Stabilizer' effect in Adobe After Effects CC 16.0 which is based on the partial 3D `subspace' stabilization method \cite{liu2011subspace}. 

\begin{figure*}[!t] 
\centering
\begin{subfigure}{\linewidth}
\includegraphics[width=\textwidth]{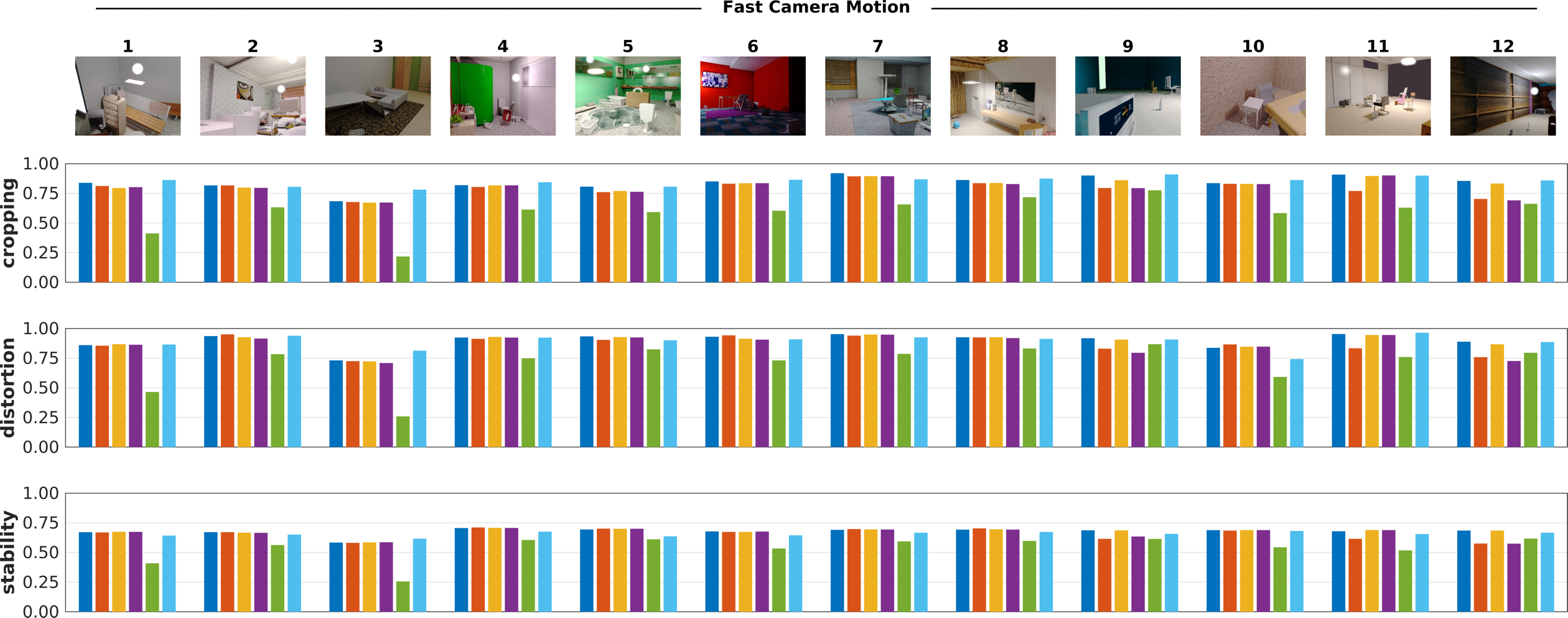}
\end{subfigure}
\vskip \baselineskip
\begin{subfigure}{\linewidth}
\centering
\includegraphics[width=.7\textwidth]{Figures/eval_legend.png}
\end{subfigure}
\caption{Quantitative comparisons with synthetic videos from the SceneNet RGB-D dataset \cite{McCormac:etal:ICCV2017}.}
\label{fig:eval_scene}
\end{figure*}

To perform comparisons, we selected 12 videos from the TUM RGB-D dataset and group them into three categories based on their content: (1) Simple, (2) Fast Camera Motion, and (3) Dynamic Objects. We then randomly select 20 different 30 frame subsequences from each video for evaluation. Similarly, we selected 12 synthetic videos from the SceneNet RGB-D dataset and selected 20 subsequences from each. Videos from this dataset are sampled at approximately 1 fps and as a result,  consist exclusively of fast camera motions, many of which we consider to be highly challenging. As such, we limited the length of the subsequences to 20 frames. We accounted for failure cases (which we defined as instances in which the stabilizing transformations map at least one input frame entirely out of view of the image plane) by setting the values of all metrics to zero for the subsequence. The average performance of each method across all subsequences from each dataset is shown in Figure \ref{fig:eval_slam} and Figure \ref{fig:eval_scene}. 

All four implementations of QuICK consistently produce high-quality stabilization. In particular, videos stabilized via QuICK suffer from less distortion and cropping than those stabilized using either the `depth camera' method or Adobe's `Warp Stabilizer' while providing comparable stability. Moreover, QuICK performs significantly better than the `depth camera' method on both real and synthetic videos containing fast camera motions. This is likely a result QuICK's reliance on dense scene flow for motion estimation instead of the feature-based approach used by the 'depth camera' method. In particular, many of the synthetic sequences from the SceneNet dataset contain rapid rotations that result in too few or non-existent matches between adjacent frames, causing  the `depth camera' method to fail. In contrast,  scene flow estimation remains intact in these instances, allowing QuICK to recover the camera motion.  While scene flow estimation becomes less accurate under large displacements, we find that our formulation of the optimal camera path is very robust, enabling QuICK to well-stabilize challenging sequences as long as the magnitude of the estimated flow remains relatively consistent.  

Our implementations of QuICK do not explicitly handle dynamic scene content. When large, near-range dynamic occlusions occur in the scene,  such as a when a person walks directly in front of the camera, the synthesized view tends to `follow' the dynamic object instead of focusing on the obvious point of interest in the scene. To minimize this effect, a technique similar to the one presented in \cite{liu2014steadyflow} could be implemented to preprocess the scene flow before motion estimation by detecting and removing sections of the flow corresponding to dynamic content.

\section{Conclusion}
Our pivotal contribution is QuICK --- a framework for robust, full 3D video stabilization that relies \textit{only} on dense scene flow to recover 3D camera motion. In particular, QuICK is based on a novel camera motion model through which camera poses can be recovered directly from 3D flows. This model can be generalized to describe non-rigid video artifacts of the type often found in consumer videos, such as those produced by zooms. Combined with efficient flow calculations, its simplicity makes QuICK fast. By circumventing the use of image features, QuICK is able to produce the kind of high-quality stabilization achieved by prior 3D methods without fragility the inherent in feature-based approaches. 

\section{Acknowledgements}
This work was performed under National Science Foundation grant IIS-1619050 and Office of Naval Research Award N00014-17-1-2142.

{\small
\bibliographystyle{IEEEtran}
\bibliography{QuICK_Ref}

\begin{thebibliography}{10}
\providecommand{\url}[1]{#1}
\csname url@samestyle\endcsname
\providecommand{\newblock}{\relax}
\providecommand{\bibinfo}[2]{#2}
\providecommand{\BIBentrySTDinterwordspacing}{\spaceskip=0pt\relax}
\providecommand{\BIBentryALTinterwordstretchfactor}{4}
\providecommand{\BIBentryALTinterwordspacing}{\spaceskip=\fontdimen2\font plus
\BIBentryALTinterwordstretchfactor\fontdimen3\font minus
  \fontdimen4\font\relax}
\providecommand{\BIBforeignlanguage}[2]{{%
\expandafter\ifx\csname l@#1\endcsname\relax
\typeout{** WARNING: IEEEtran.bst: No hyphenation pattern has been}%
\typeout{** loaded for the language `#1'. Using the pattern for}%
\typeout{** the default language instead.}%
\else
\language=\csname l@#1\endcsname
\fi
#2}}
\providecommand{\BIBdecl}{\relax}
\BIBdecl

\bibitem{grundmann2011auto}
M.~Grundmann, V.~Kwatra, and I.~Essa, ``Auto-directed video stabilization with
  robust {L1} optimal camera paths,'' in \emph{Proceedings of the IEEE
  Conference on Computer Vision and Pattern Recognition}.\hskip 1em plus 0.5em
  minus 0.4em\relax IEEE, 2011, pp. 225--232.

\bibitem{liu2009content}
F.~Liu, M.~Gleicher, H.~Jin, and A.~Agarwala, ``Content-preserving warps for 3d
  video stabilization,'' \emph{ACM Transactions on Graphics}, vol.~28, no.~3,
  p.~44, 2009.

\bibitem{liu2011subspace}
F.~Liu, M.~Gleicher, J.~Wang, H.~Jin, and A.~Agarwala, ``Subspace video
  stabilization,'' \emph{ACM Transactions on Graphics}, vol.~30, no.~1, p.~4,
  2011.

\bibitem{matsushita2006full}
Y.~Matsushita, E.~Ofek, W.~Ge, X.~Tang, and H.-Y. Shum, ``Full-frame video
  stabilization with motion inpainting,'' \emph{IEEE Transactions on Pattern
  Analysis and Machine Intelligence}, vol.~28, no.~7, pp. 1150--1163, 2006.

\bibitem{liu2012video}
S.~Liu, Y.~Wang, L.~Yuan, J.~Bu, P.~Tan, and J.~Sun, ``Video stabilization with
  a depth camera,'' in \emph{Proceedings of the IEEE Conference on Computer
  Vision and Pattern Recognition}.\hskip 1em plus 0.5em minus 0.4em\relax IEEE,
  2012, pp. 89--95.

\bibitem{liu2013bundled}
S.~Liu, L.~Yuan, P.~Tan, and J.~Sun, ``Bundled camera paths for video
  stabilization,'' \emph{ACM Transactions on Graphics}, vol.~32, no.~4, p.~78,
  2013.

\bibitem{sand2008particle}
P.~Sand and S.~Teller, ``Particle video: {L}ong-range motion estimation using
  point trajectories,'' \emph{International Journal of Computer Vision},
  vol.~80, no.~1, p.~72, 2008.

\bibitem{brox2010object}
T.~Brox and J.~Malik, ``Object segmentation by long term analysis of point
  trajectories,'' in \emph{Proceedings of the European Conference on Computer
  Vision}.\hskip 1em plus 0.5em minus 0.4em\relax Springer, 2010, pp. 282--295.

\bibitem{sundaram2010dense}
N.~Sundaram, T.~Brox, and K.~Keutzer, ``Dense point trajectories by
  {GPU}-accelerated large displacement optical flow,'' in \emph{Proceedings of
  the European Conference on Computer Vision}.\hskip 1em plus 0.5em minus
  0.4em\relax Springer, 2010, pp. 438--451.

\bibitem{wang2013dense}
H.~Wang, A.~Kl{\"a}ser, C.~Schmid, and C.-L. Liu, ``Dense trajectories and
  motion boundary descriptors for action recognition,'' \emph{International
  Journal of Computer Vision}, vol. 103, no.~1, pp. 60--79, 2013.

\bibitem{liu2014steadyflow}
S.~Liu, L.~Yuan, P.~Tan, and J.~Sun, ``Steadyflow: {S}patially smooth optical
  flow for video stabilization,'' in \emph{Proceedings of the IEEE Conference
  on Computer Vision and Pattern Recognition}, 2014, pp. 4209--4216.

\bibitem{liu2016meshflow}
S.~Liu, P.~Tan, L.~Yuan, J.~Sun, and B.~Zeng, ``Meshflow: {M}inimum latency
  online video stabilization,'' in \emph{Proceedings of the European Conference
  on Computer Vision}.\hskip 1em plus 0.5em minus 0.4em\relax Springer, 2016,
  pp. 800--815.

\bibitem{guo2018view}
H.~Guo, S.~Liu, S.~Zhu, H.~T. Shen, and B.~Zeng, ``View-consistent {MeshFlow}
  for stereoscopic video stabilization,'' \emph{IEEE Transactions on
  Computational Imaging}, vol.~4, no.~4, pp. 573--584, 2018.

\bibitem{jaimez2015primal}
M.~Jaimez, M.~Souiai, J.~Gonzalez-Jimenez, and D.~Cremers, ``A primal-dual
  framework for real-time dense {RGB-D} scene flow,'' in \emph{Proceedings of
  the IEEE International Conference on Robotics and Automation}.\hskip 1em plus
  0.5em minus 0.4em\relax IEEE, 2015, pp. 98--104.

\bibitem{shi1993good}
J.~Shi and C.~Tomasi, ``Good features to track,'' in \emph{Proceedings of the
  IEEE Conference on Computer Vision and Pattern Recognition}.\hskip 1em plus
  0.5em minus 0.4em\relax IEEE, 1994, pp. 593--600.

\bibitem{gleicher2007re}
M.~L. Gleicher and F.~Liu, ``Re-cinematography: {I}mproving the camera dynamics
  of casual video,'' in \emph{Proceedings of the ACM International Conference
  on Multimedia}.\hskip 1em plus 0.5em minus 0.4em\relax ACM, 2007, pp. 27--36.

\bibitem{goldstein2012video}
A.~Goldstein and R.~Fattal, ``Video stabilization using epipolar geometry,''
  \emph{ACM Transactions on Graphics}, vol.~31, no.~5, p. 126, 2012.

\bibitem{wang2013spatially}
Y.-S. Wang, F.~Liu, P.-S. Hsu, and T.-Y. Lee, ``Spatially and temporally
  optimized video stabilization,'' \emph{IEEE Transactions on Visualization and
  Computer Graphics}, vol.~19, no.~8, pp. 1354--1361, 2013.

\bibitem{liu2017hybrid}
S.~Liu, B.~Xu, C.~Deng, S.~Zhu, B.~Zeng, and M.~Gabbouj, ``A hybrid approach
  for near-range video stabilization,'' \emph{IEEE Transactions on Circuits and
  Systems for Video Technology}, vol.~27, no.~9, pp. 1922--1933, 2017.

\bibitem{hartley2003multiple}
R.~Hartley and A.~Zisserman, \emph{Multiple View Geometry in Computer Vision},
  2nd~ed.\hskip 1em plus 0.5em minus 0.4em\relax New York, NY, USA: Cambridge
  University Press, 2003.

\bibitem{smith2009light}
B.~M. Smith, L.~Zhang, H.~Jin, and A.~Agarwala, ``Light field video
  stabilization,'' in \emph{Proceedings of the IEEE International Conference on
  Computer Vision}.\hskip 1em plus 0.5em minus 0.4em\relax IEEE, 2009, pp.
  341--348.

\bibitem{kroeger2016fast}
T.~Kroeger, R.~Timofte, D.~Dai, and L.~Van~Gool, ``Fast optical flow using
  dense inverse search,'' in \emph{Proceedings of the European Conference on
  Computer Vision}.\hskip 1em plus 0.5em minus 0.4em\relax Springer, 2016, pp.
  471--488.

\bibitem{Sun_2018_CVPR}
D.~Sun, X.~Yang, M.-Y. Liu, and J.~Kautz, ``{PWC-Net}: {CNN}s for optical flow
  using pyramid, warping, and cost volume,'' in \emph{Proceedings of the IEEE
  Conference on Computer Vision and Pattern Recognition}, June 2018.

\bibitem{taniai2017fast}
T.~Taniai, S.~N. Sinha, and Y.~Sato, ``Fast multi-frame stereo scene flow with
  motion segmentation,'' in \emph{Proceedings of the IEEE Conference on
  Computer Vision and Pattern Recognition}, 2017, pp. 3939--3948.

\bibitem{chirikjian2018pose}
G.~S. Chirikjian, R.~Mahony, S.~Ruan, and J.~Trumpf, ``Pose changes from a
  different point of view,'' \emph{Journal of Mechanisms and Robotics},
  vol.~10, no.~2, p. 021008, 2018.

\bibitem{chirikjian2011stochastic}
G.~S. Chirikjian, \emph{Stochastic Models, Information Theory, and Lie Groups,
  Volume 2: {A}nalytic Methods and Modern Applications}.\hskip 1em plus 0.5em
  minus 0.4em\relax Springer Science \& Business Media, 2011, vol.~2.

\bibitem{Murray:1994:MIR:561828}
R.~M. Murray, S.~S. Sastry, and L.~Zexiang, \emph{A Mathematical Introduction
  to Robotic Manipulation}, 1st~ed.\hskip 1em plus 0.5em minus 0.4em\relax Boca
  Raton, FL, USA: CRC Press, Inc., 1994.

\bibitem{ma2012invitation}
Y.~Ma, S.~Soatto, J.~Kosecka, and S.~S. Sastry, \emph{An {I}nvitation to 3-{D}
  {V}ision: {F}rom {I}mages to {G}eometric {M}odels}.\hskip 1em plus 0.5em
  minus 0.4em\relax SpringerVerlag, 2003.

\bibitem{zacur2014left}
E.~Zacur, M.~Bossa, and S.~Olmos, ``Left-invariant {R}iemannian geodesics on
  spatial transformation groups,'' \emph{SIAM Journal on Imaging Sciences},
  vol.~7, no.~3, pp. 1503--1557, 2014.

\bibitem{andruchow2014left}
E.~Andruchow, G.~Larotonda, L.~Recht, and A.~Varela, ``The left invariant
  metric in the general linear group,'' \emph{Journal of Geometry and Physics},
  vol.~86, pp. 241--257, 2014.

\bibitem{vandereycken2013riemannian}
B.~Vandereycken, P.-A. Absil, and S.~Vandewalle, ``A {R}iemannian geometry with
  complete geodesics for the set of positive semidefinite matrices of fixed
  rank,'' \emph{IMA Journal of Numerical Analysis}, vol.~33, no.~2, pp.
  481--514, 2013.

\bibitem{chirikjian2017signal}
G.~S. Chirikjian, ``Signal classification in quotient spaces via globally
  optimal variational calculus,'' in \emph{Proceedings of the IEEE Conference
  on Computer Vision and Pattern Recognition Workshops}, 2017, pp. 56--64.

\bibitem{kim2006conformational}
J.~S. Kim and G.~S. Chirikjian, ``Conformational analysis of stiff chiral
  polymers with end-constraints,'' \emph{Molecular Simulation}, vol.~32,
  no.~14, pp. 1139--1154, 2006.

\bibitem{han2001least}
Y.~Han and F.~C. Park, ``Least squares tracking on the {E}uclidean group,''
  \emph{IEEE Transactions on Automatic Control}, vol.~46, no.~7, pp.
  1127--1132, 2001.

\bibitem{liu2017codingflow}
S.~Liu, M.~Li, S.~Zhu, and B.~Zeng, ``{CodingFlow}: {E}nable video coding for
  video stabilization,'' \emph{IEEE Transactions on Image Processing}, vol.~26,
  no.~7, pp. 3291--3302, 2017.

\bibitem{sturm12iros}
J.~Sturm, N.~Engelhard, F.~Endres, W.~Burgard, and D.~Cremers, ``A benchmark
  for the evaluation of {RGB-D SLAM} systems,'' in \emph{Proceedings of the
  International Conference on Intelligent Robot Systems}, Oct. 2012.

\bibitem{McCormac:etal:ICCV2017}
J.~McCormac, A.~Handa, S.~Leutenegger, and A.~J. Davison, ``{SceneNet RGB-D}:
  {C}an {5M} synthetic images beat generic {ImageNet} pre-training on indoor
  segmentation?'' in \emph{Proceedings of the IEEE International Conference on
  Computer Vision}, 2017.

\end{thebibliography}
}

\end{document}